\documentclass{article} 
\usepackage{graphicx}
\usepackage{subfigure}
\usepackage{url}   
\usepackage{lipsum}
\usepackage{booktabs}       
\usepackage{amsfonts}       
\usepackage{nicefrac}       
\usepackage{xcolor}         
\usepackage{amsmath}
\usepackage{array}
\usepackage{siunitx}
\usepackage{multirow}
\usepackage{nicematrix}
\usepackage{adjustbox}
\DeclareMathOperator{\EX}{\mathbb{E}}
\usepackage{hyperref}
\usepackage[utf8]{inputenc} 
\usepackage[T1]{fontenc}    
\usepackage{microtype}      

\begin{document}

\title{Global-Local  Processing in \\Convolutional Neural Networks}

\author{Zahra Rezvani*$^a$, Soroor Shekarizeh$^a$, Mohammad Sabokrou$^{a,b}$ \\ \\
$^a$Institute for Research in Fundamental Sciences, Tehran, Iran \\
$^b$Okinawa Institute of Science and Technology, Okinawa, Japan
}

\maketitle
\begin{abstract}

Convolutional Neural Networks (CNNs) have achieved outstanding performance on image processing challenges. Actually, CNNs  imitate the typically developed human brain structures at the micro-level (Artificial neurons). At the same time,  they distance themselves from imitating natural visual perception in humans at the macro architectures (high-level cognition). Recently it has been investigated that CNNs are highly biased toward local features and fail to detect the global aspects of their input. Nevertheless,  the literature offers limited clues on this problem. To this end, we propose a simple yet effective solution inspired by the unconscious behavior of the human pupil.  We devise a simple module called Global Advantage Stream (GAS) ($\mathcal{G}$) to learn and capture the holistic features of input samples (i.e., the global features). Then, the GAS features were combined with a CNN network as a plug-and-play component called the Global/Local Processing (GLP) model. The experimental results confirm that this stream improves the accuracy with an insignificant additional computational/temporal load and makes the network more robust to adversarial attacks. Furthermore, investigating the interpretation of the model shows that it learns a more holistic representation similar to the perceptual system of healthy humans \footnote{Source code is available \href{https://github.com/rezvanizahra/GLP.git}{here}}.

\end{abstract}

\section{Introduction}
\label{Introduction}
Deep learning methods, as a cutting edge of artificial intelligence, are trained by filtering information through multiple hidden layers. 
The current DNNs can mimic the human brain at the micro-level (Neuronal level) but fails to deal with  macro-level network behavior( Cognitive level). 

\begin{figure}[h]
    \centering
    \includegraphics[width=\textwidth]{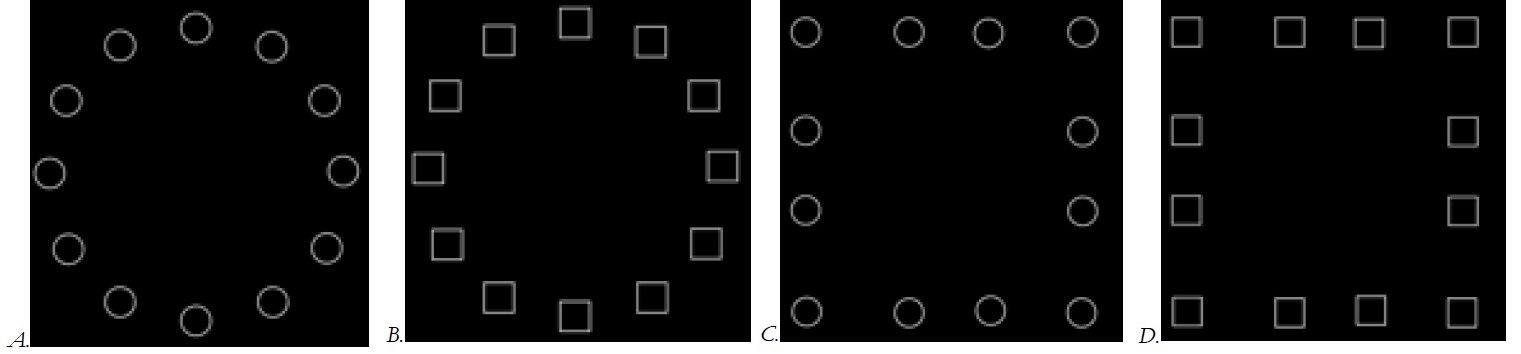}
    \caption{Examples of hierarchical compound stimuli, that are commonly used  to evaluate the ability to detect at the general and partial levels separately in diagnostic applications. They were used to design the global advantage layer in experiment 1.  }
    \label{fig:Navon}
\end{figure}

 \cite{baker2018deep} shows that deep convolutional neural networks have more tendency to use texture information over the general shape. They have tried to train CNNs to categorize images using artificial images with misleading textures and suggest that texture plays a vital role in CNNs. They claimed that deep learning systems have no sensitivity to the overall shape of images and show that benchmark CNNs can not distinguish bounding contours of objects.
While it is investigated that humans attend to the features of the general shape before considering the local features such as texture \cite{hermann2020origins}.
The stimuli in the natural world have inherently hierarchical architecture: general form or global level and detail texture or local level. 
Global/Local processing (GLP) is one of the important early debates in psychology about the human perceptual system throughout the past four decades and is still an ongoing challenge. Global Precedence Effect (GPE), as a modern version of Gestalt theory, claims that individuals more readily, process global features faster than local details \cite{navon1977forest}.
GPE has been investigated in a series of experiments with hierarchical compound stimuli consisting of a global letter/shape formed by the configuration of local letters/shapes (See Fig.~ \ref{fig:Navon}), which are independent in local and global levels of the stimuli. This phenomenon is responsible for the ability to generalize in humans. So, it helps us perceive the forest before the trees and categorize them correctly, despite the differences in the details of the objects\cite{navon1977forest}.

Pupil diameter is subconsciously controlled by the brain due to environmental stimuli. As the pupil shrinks, the reflection of images is focused on the fovea (located in the center of the Retina), which is made up mostly of Cones photoreceptor cells. These cells are responsible for receiving high-level features or details. Instead, the area around the Fovea is filled with Rodes cells. These cells are responsible for low-level features and have a low spatial acuity. As the pupil opens, the reflection of the environment is received by both cell types.
The pupil is primarily regulated by prevailing light levels but is also modulated by perceptual and attentional factors. \cite{sabatino2018task} found through psychophysical experiments with hierarchical stimuli that individuals have a characteristic constriction of the pupil waveform during the selection of local information relative to global information. They indicate that pupil changes may serve as a visual filtering mechanism important for attentional selection. This work represented the first characterization of pupil response in the context of selective attention and suggested that mechanisms underlying the earliest stages of visual processes could be relevant for perception and visual selection. 
Also, it has been observed that children with ASD showed pupil constriction as a response to  images  of faces  \cite{anderson2006visual}. However, neurotypical  children showed pupil dilation in response to the same stimuli \cite{de2021autism}. There is other evidence that pupillometry reliably tracks inter-individual differences in perceptual style as a biomarker, and individuals with typically developed perception distribute attention to both surfaces in a more global, holistic style \cite{turi2018pupillometry}. 
Recently it has been investigated that Vision Transformer(ViT) has a better performance on modeling the holistic features of images\cite{dosovitskiy2020image}. They  split the images into fixed-size patches, embedding them, and feeding them to a Transformer Encoder (TE) \cite{dosovitskiy2020image}.TE was inspired by vanilla transformers introduced for NLP tasks \cite{vaswani2017attention}. \cite{aldahdooh2021reveal} assessed such  models as more robust to adversarial examples. Nevertheless, the ViT models  are computationally expensive. 

In this paper, Global Advantage Stream (GAS) was added to increase the accuracy and robustness of common CNNs.  The purpose of this stream is to provide a holistic view of these networks, which not only increased their accuracy in categorizing images but also their resistance to common attacks dramatically improved. 
The novelty of this study is that, unlike previous state-of-the-art research, the design of GAS is completely inspired by the subconscious function of the human pupils.  Also, The function of this stream is very similar to early therapeutic intervention methods using robots in educating autistic children to facilitate decoding of the overall features of the perceptual environment by removing details and helping them to return back to the right track base on GPE in normal individuals. 
The main contributions of this paper are: \begin{itemize}
\item The presented model, unlike CNNs, can consider global features in addition to local ones. The method inspired by the subconscious function of the human pupil extracts both sets of features simultaneously. Feature sets were concatenated to classify the images based on both global form and detail texture. The existence of global features in the feature bank empowers the model to follow the top-down attention strategy in addition to the bottom-up attention approach.  
\item It has been shown that the proposed method is both more accurate and more robust than CNNs. However, the proposed model imposes an insignificant computational time load on the CNN model. Also, the proposed method has better interpretability rather than CNNs.  It has been shown that the proposed model has a better performance. Also, because of its holistic view, it is more resistant to common adversarial attacks. Furthermore, better explainability according to the XAI method confirms better localization of the whole object in the images instead of focusing on a local detailed part.
\end{itemize}

\section{Method}
The main objective of this paper is, inspired by human behavior unconsciousness,  forcing  the deep neural network to learn both global and local representation. This makes the model more accurate and robust. The new model called the Global/Local Processing (GLP) model is composed of two main components (i.e., streams) that are concatenated in parallel: (1)
The local stream ($\mathcal{L}$) and (2) The quick global stream ($\mathcal{G}$).   $\mathcal{L}$ is a conventional CNN that inherently learns the local features and complex local patterns. To competence the inability  of such models to learn the holistic features, we introduce the $\mathcal{G}$ stream.  $\mathcal{G}$ is composed of the GAS module, which is described below.  In short, this module is made of a smart filter followed by two convolutional layers (feature maps).  GAS is responsible for capturing global features inspiring  the subconscious function of the human pupil. 
\begin{figure}[ht]
    \centering   \includegraphics[width=\textwidth]{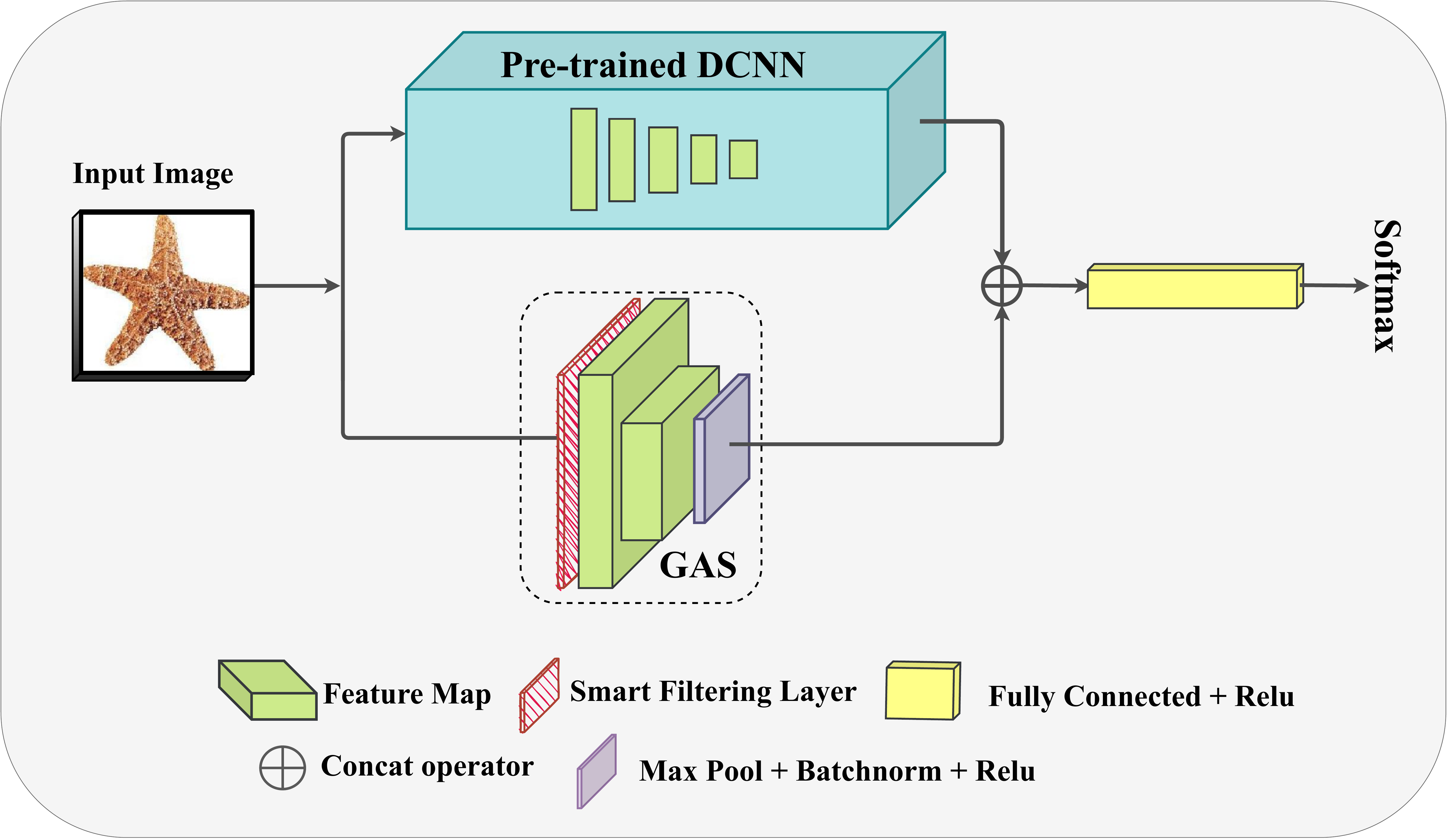}
    \caption{Overview of the proposed method (GLP model). Global and local features are extracted through separated streams and then all the features concatenate to classify the images.}
    \label{fig:model_arc}
\end{figure}
Fig.~ \ref{fig:model_arc} shows the overall schema of the proposed method. The inside information of $\mathcal{G}$ and $\mathcal{L}$ components and training procedure is explained in  the following subsections.

\subsection{GAS: Global Advantage Stream}
\begin{figure}[ht]
    \centering
      \subfigure[The Fovea and the distribution of Rods and Cones in the human eye.]{\includegraphics[width=0.45\textwidth]{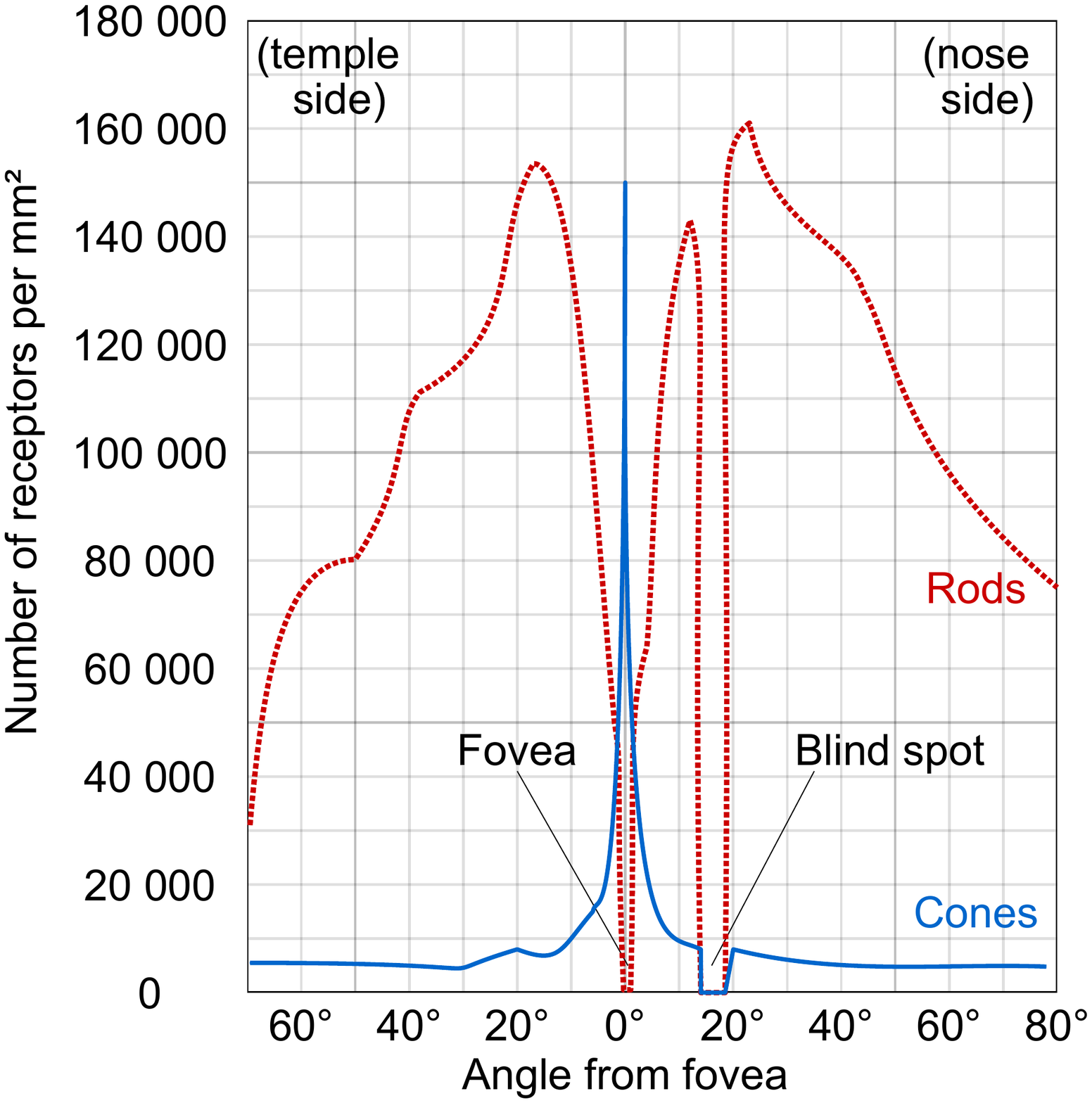}}
    \subfigure[Smart filtering visualisation]{\includegraphics[width=0.45\textwidth]{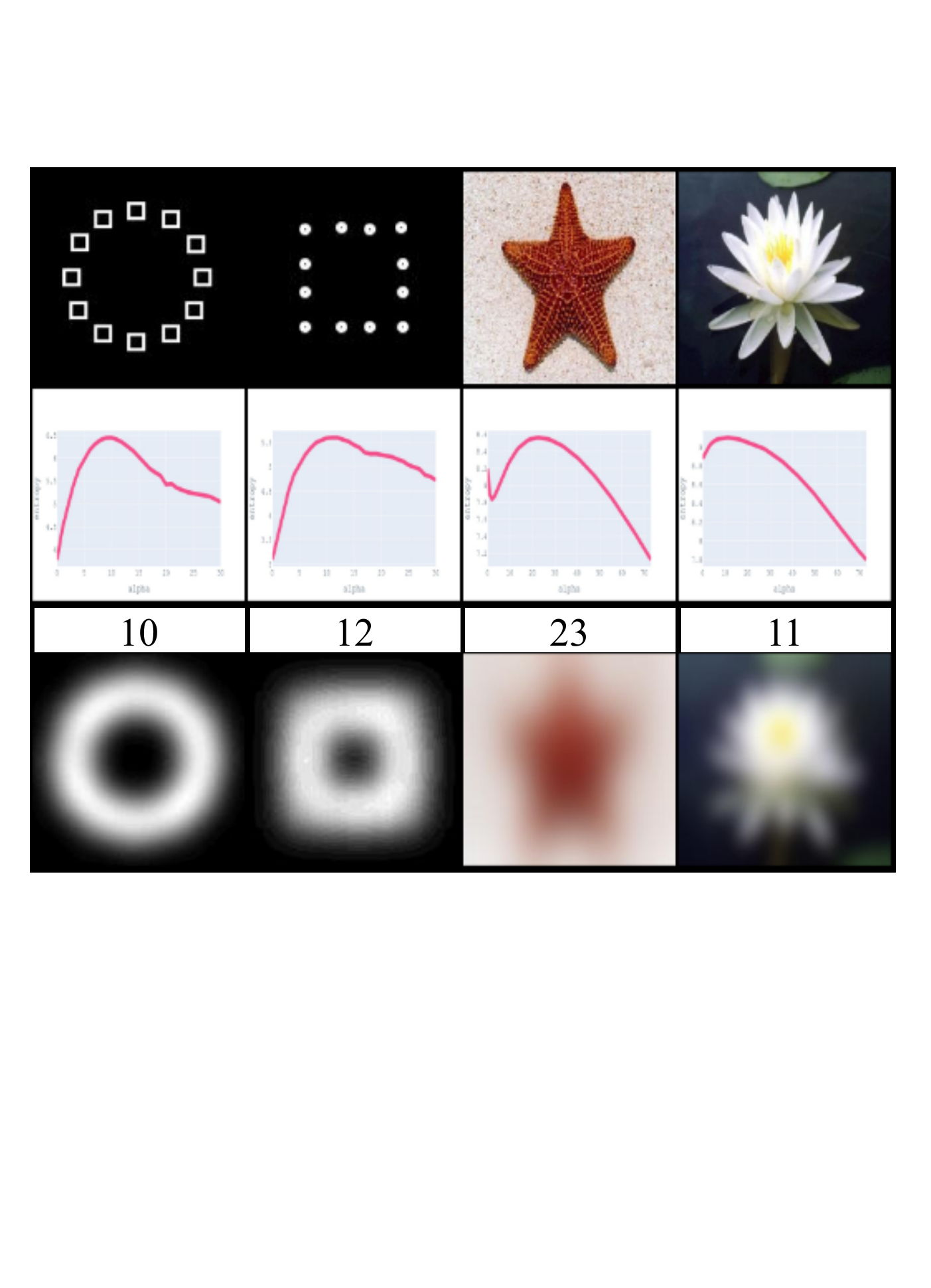}} 
    \caption{ a) Constricted pupil reflexes all the ambient light in Fovea covered by Cones receptors with high acuity. But when the pupil dilates, image reflexes are received simultaneously by Rodes receptors with low acuity based on the cells' distribution in the human eyes. Rodes cells are more responsible for peripheral vision \cite{wandell1995foundations}.  b) Examples of Smart filtering visualization.  \textbf{First column:} input images, \textbf{Second column:} diagrams of changes in the amount of entropy by increasing the $\alpha$, \textbf{Third column: } the optimum  $\alpha$ that maximizes the value of entropy in the filtered image, and \textbf{last column:} outputs of applying the proposed smart filtering using the optimum  $\alpha$.}
    \label{fig:SmartFiltering}
\end{figure}

The goal of GAS is to extract global features. In fact, this stream is inspired by the subconscious function of the human pupil. During focus, the environment projects only on Fovea (populated exclusively by Cones with high spatial acuity), but in normal situations with dilated pupils, most of the ambient light is received by  Rodes  with low spatial acuity. Fig.~  \ref{fig:SmartFiltering} (a) displays the frequency distribution of these two cell types relative to the distance from the Fovea.

 In the GAS layer, firstly, a smart low-pass filter in the frequency domain has applied to the image input aiming to attenuate high-frequency noises. The most important point about the layer is the automatic cut-off parameter ($\alpha$) setting according to each input. To find the proper value for alpha smartly, we used the Entropy criterion. The amount of uncertainty in an entire probability distribution is quantified using the Shannon entropy.
The entropy is calculated from the following Equ.~\ref{eq:EntropyFormul}.$\mathcal{I(X)}$ is defined as self-information of an event $\mathcal{X}=x$ ~~\cite{goodfellow2016deep}. It is investigated that by increasing the radius of the Gaussian low-pass filter, the image entropy also will slightly increase. But after this phase, the routine continues in reverse. With increasing this parameter, the entropy changes in the opposite direction. According to this finding, as a next step, we find the value of $\alpha$ that maximize the Entropy of the input (filtered image) based on Equ.~\ref{eq:OptimumAlpha}. Interestingly, by smart filtering selectively with this value, all the local information has faded, and instead, the global structure of the image is more readily detected. The value of the optimum $\alpha$ varies based on each image structure and size (see Fig.~\ref{fig:SmartFiltering} (b)).
In GAS, after removing the local details smartly, there are two feature map layers followed by two layers  for pooling and batch normalization( see Fig.~ \ref{fig:model_arc} ). 

\begin{equation}
      {H}(x) =\EX_{x \sim P}[\mathcal{I(X)}] = -\EX_{x \sim P}[P(x)]
      \label{eq:EntropyFormul}
\end{equation}

\begin{equation}
      \alpha ^{*} = \arg \max _{\alpha}H(filteredImage(X,\alpha)) 
      \label{eq:OptimumAlpha}
\end{equation}

\subsection{Global/Local processing Model}

The GLP Model is designed in such a way that global features are obtained from one stream, and local features are achieved from another stream. Afterward, they concatenated with each other as an ultimate feature bank. Finally, there is a fully connected layer to map the features to the output category. As shown in Fig.~ \ref{fig:model_arc}, the local stream  consists of a pre-trained CNN. As we discussed before in the Introduction section \ref{Introduction}, CNNs extract features based on local details in images. So, most of them could be considered as the local stream ($\mathcal{L}$) in our model.  On the other hand, the global stream ($\mathcal{G}$) is made up of a GAS module.

\textbf{$\mathcal{G+L}$ Training:} The training of this hybrid model is done in several steps as follows:
\begin{enumerate}
\item For $\mathcal{L}$ layer weights, we use the pre-trained weights of common  CNN models, and local features are calculated using them.
\item The GAS module is trained on training data to extract global features.
\item Then, the two feature sets are concatenated, and the feature bank has been completed.
\item Finally, the fully connected layer is trained so that it can perform the classification with the best accuracy.
\end{enumerate}

In the following section, we will show how GAS works to extract global features  in the first experiment \ref{exp:epx1}. Then we represent how this new stream improves the accuracy and robustness of the benchmark models in the next section in the second experiment \ref{exp:epx2}. After that, we demonstrate that the interpretability of the model increases via an Explainable AI method (see Fig.~ \ref{fig:visualization}). 

\section{Results} 
To evaluate the proposed method, we have designed two experiments. Firstly, we investigated how GAS extracts global shape using the Navon dataset. Also, We showed that the common-used CNNs couldn't succeed in this simple task. Then, using an XAI algorithm presented how the GAS extracts exactly the global shape, but the others only focus on local details. 

We also evaluated the GLP model accuracy using the Caltech101 dataset \cite{FeiFei2004LearningGV}. Furthermore, we evaluated the robustness of our model  against adversarial attacks. Moreover, to demonstrate the interpretability of our method, visualizations of feature maps were presented.

We have exploited the Gradient-weighted Class Activation Mapping (Grad-CAM) \cite{selvaraju2017grad}  to showcase the interpretability of the models in both experiments. Grad-CAM highlights the most important areas in the image by making the gradient of the classification score regarding the final convolutional layer.

All experiments are implemented using PyTorch \cite{paszke2019pytorch} and performed using an NVIDIA GeForce RTX 2060 GPU.


\subsection{GAS module design and comparison}
\label{exp:epx1}

In this experiment, we have compared the performance of the GAS model with two common-use CNN networks, Resnet18 \cite{he2016deep} and InceptionV3 \cite{szegedy2015going} to classify  the Navon compound stimuli dataset  \cite{navon1977forest} based on the global/local shape. We trained the GAS network and fine-tuned pre-trained Resnet18 and Inception-v3 models with simple augmented shape images (3000 images in two categories) to  recognize the shape of circles and squares. Then, we evaluate all the networks with the Navon compound stimuli dataset based on the local/ global level. The initial learning rate for training the GAS network and fine-tuning the pre-trained models is equal to $2e^{-3}$ with a decay rate. Also, we used stochastic gradient descent (SGD) optimizer with a momentum equal to 0.9 and a batch size of 64 for all the models.
All the information about the computational and time load has been summarized in Appendix \ref{model_performance}.

\textbf{Navon Dataset} \cite{navon1977forest} Navon is a set of compound stimuli with independent information at the local vs. global level (see \ref{fig:Navon}). This experiment used geometrical shapes (circle vs. square) for both levels in different sizes, sparsity, and line width (4152 hierarchical images). The dataset has been used to test the ability to detect global/local shape detection. The Navon dataset and dataset of simple shape images have been included in the supplementary material.

\textbf{Results}.  The results of this experiment have been summarized in Table \ref{top1_Navon}. This listed the top 1  model accuracy on the Navon dataset based on both Local / Global image shapes. As illustrated, the GAS outperforms two other CNNs in Global shape detection. While Resnet18 and InceptionV3 obtain better performance for local shape detection. 

\begin{table}
  \caption{Top1 Accuracy (\%) of the models in global/local shape detection tasks on Navon dataset}
  \label{top1_Navon}
  \centering
  \begin{tabular}{lll}
    \toprule
    Model  & Acc on Local & Acc on Global  \\
    \midrule
    Resnet18  & \textbf{85.24} &	53.65  \\
    \midrule
    InceptionV3 & \textbf{75.6 } &	62.15  \\
     \midrule
    Global Advantage Stream  &  56.42 &	\textbf{86.28}  \\
    \bottomrule
  \end{tabular}
\end{table}

\begin{figure}[ht]
    \centering
\includegraphics[width=0.8\linewidth]{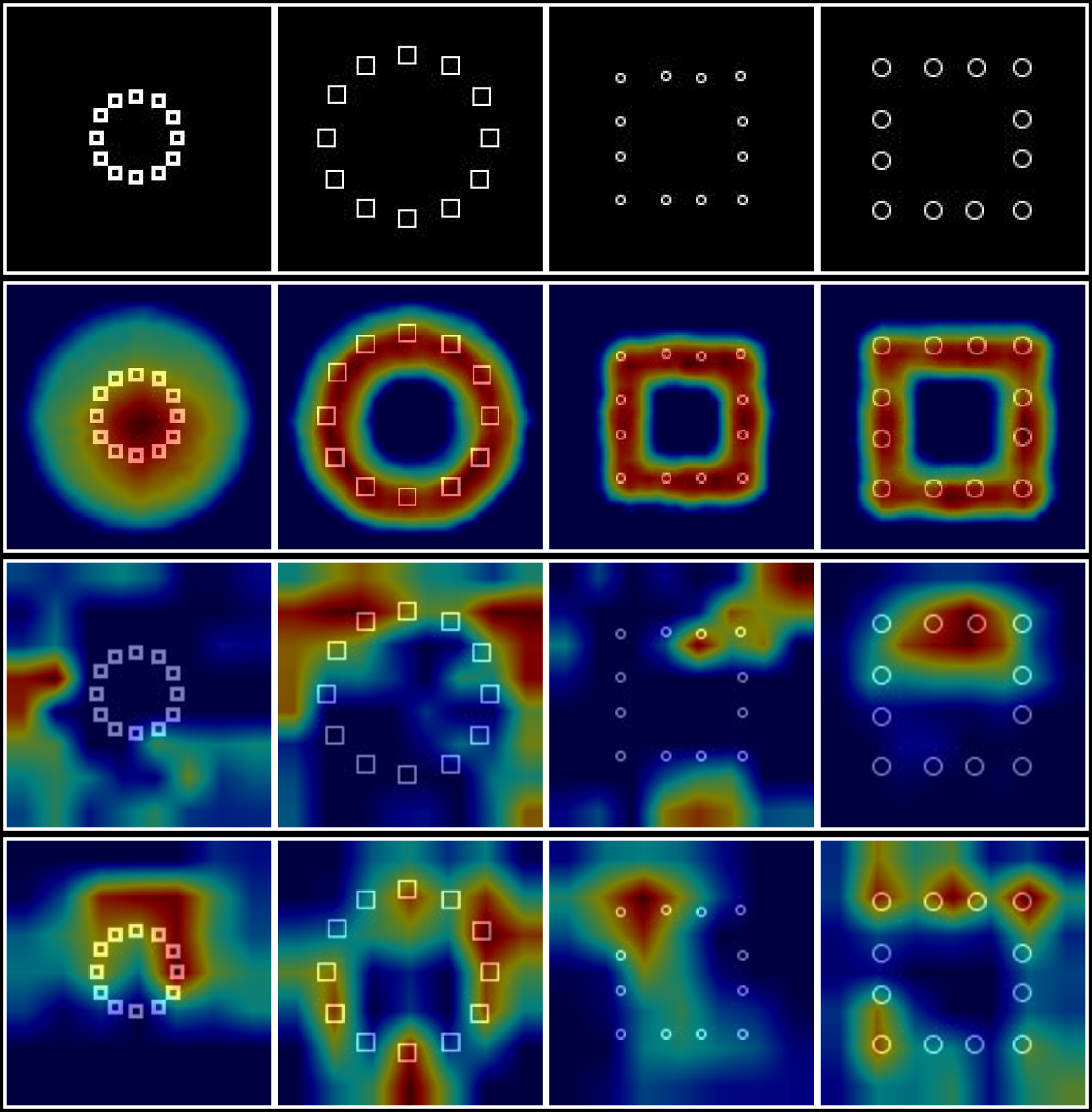}
    \caption{Comparison of visualization on the Navon dataset global test images. The top row shows input images, and the rest of the rows depict the visualization results of GAS, InceptionV3, and Resnet18, respectively }
    \label{fig:novan_visualization}
\end{figure}

\textbf{Visualization}. Fig.~ \ref{fig:novan_visualization} represented the visualized feature maps in the global shape detection task.  These tables confirm the power of GAS as a global feature extractor. This is while the CNN models failed to localize the global shape and only heat the local features.

\textbf{Discussion}.
In this experiment, we confirm that 
\begin{enumerate}
    \item The commonly used heavy CNNs behave weakly in global form detection instead more powerfully in local shape detection and detail texture extraction. 
    \item The GAS module, despite the higher speed (according to Appendix \ref{app_performance}), is more powerful in extracting global features inspires by the subconscious function of the human pupil.
    \item Visualizations confirm the global detection ability in GAS, unlike the other CNNs. 
\end{enumerate}

In the next experiment, we  improved the CNNs by combining GAS module as an extra parallel  stream to them. Then evaluate the new hybrid model in  accuracy and robustness. 

\subsection{GLP model evaluation and explainability}
\label{exp:epx2}
In the second experiment, we aimed to evaluate the GLP model's classification accuracy and its robustness in facing adversarial attacks compared to the mentioned CNNs on the Caltech101 dataset. For comparison, we train a GLP model (called GA-Resnet) using a pre-trained Resnet18 as the $\mathcal{L}$ stream, and our pre-trained GAS model (GAS-299) on the Caltech101 dataset as the  $\mathcal{G}$ stream, also in the same way for comparing our GLP model (called GA-Inception) with the pre-trained InceptionV3. 
Similar to the previous experiment, we trained the GLP network and fine-tuned pre-trained Resnet18 and Inception-v3 models by the initial Learning rate $2e^{-3}$ with a decay rate, the SGD optimizer, a batch size of 8, and with the standard input images size $299\times299$ for InceptionV3, GLP-299, and GA-Inception models, also $224\times224$ for Resnet18, GLP-224, and GA-Resnet models. 
 
\textbf{Caltech101} \cite{FeiFei2004LearningGV} is a well-known dataset for object classification which consists of $\sim 9K$ images belonging to 101 classes (e.g., “starfish”, “dolphin” and “umbrella” etc.)  and a background clutter class that contains different objects from the 101 categories. 
To evaluate our approach, we did not use the background images and split the rest of the 101 classes of images into the train(60\%), validation(20\%), and test(20\%).  

\textbf{Adversarial Attacks}
For evaluating robustness we apply two common adversarial attacks called Fast Gradient Sign Method (FGSM) \cite{goodfellow2014explaining} and Projected Gradient Descent (PGD) \cite{kurakin2018adversarial}. To employ attacks, we used the CleverHans \cite{papernot2016technical} which is a Python library for adversarial attacks. Both FGSM and PGD are categorized in White box attacks which means that the attacker has access to the model’s parameters.

FGSM attack, which first is introduced by \cite{goodfellow2014explaining} is a simple yet effective method by using the gradients of a CNN to generate adversarial images. As Equ.~ \ref{eq:fgsm} is defined, for an input image $x$, FGSM computes the loss of the model prediction regarding the actual class label, then calculates the gradients of the loss with respect to the input image, and uses the sign of the gradients to create the new adversarial image $Adv(x)$ which maximizes the loss. For a given input image $x$, the adversarial image is generated as follows: 

\begin{equation}
      Adv(x) = x+ \epsilon * sing( \bigtriangledown_x \mathcal{J}(\theta, x, y))
      \label{eq:fgsm}
\end{equation}
where $y$ is the input actual label, $\epsilon$ is used to ensure the perturbations are small enough to can not detect by human eyes but large enough to fool the CNN, $\mathcal{J}$ is the model loss function, and $\theta$ is the model parameters.

PGD attack generates new adversarial images in an iterative scheme. Following Equ.~ \ref{eq:pgd}, PGD tries to maximize the loss of the CNN model on an input image $x$ while finding a perturbation smaller than $\epsilon$. Besides defining $\epsilon$ as the maximum perturbation size, it is required to determine a metric to calculate the distance from the adversarial image $Adv(x)$ to the input image $x$. This metric ensures that the output adversarial example is not perceptibly different from humans. Among the various
$L_p$ norms  $p = 2$ or $p = \infty$ are the most common-use \cite{carlini2017towards}. The PGD  attack is formulated as
follows: 

\begin{equation}
      Adv(x)_i = CLIP_{x,\epsilon}(x_{i-1} + \lambda sing \bigtriangledown_x(\mathcal{J}(.)));    Adv(x)_0 =x_{original}, 
      \label{eq:pgd}
\end{equation}
where $i$ defined the number of iterations, $CLIP$ is an operation that clips $x$ back to the permissible set, $\lambda$ is the step size and  $\mathcal{J}$ is the model loss function.

To evaluate the robustness of the GLP model, we investigated the accuracy changes by increasing the perturbation size for the FGSM attack($\epsilon$) and the maximum perturbation size($\epsilon$) for the PGD attack. Following \cite{reddy2020biologically}, we reported the accuracy of our GLP model compared to the fine-tuned InceptionV3 and Resnet18 models on various $\epsilon = [0, 0.001, 0.005, 0.01, 0.05, 0.1, 0.15, 0.5]$ for both attacks. We calculated the accuracy as 1 - (naturally misclassified images + adversarial misclassified examples) since we run the adversarial attacks only on images that were not naturally misclassified. All the experiments are repeated for five iterations and the average accuracy was reported.
For the PGD attack, we report $L_{\infty}$ PGD results in our experiments.  Also,  we set the step size $\lambda$ to $\epsilon/3$ since it allows the PGD attack to reach the edge of the permissible set and explore the boundary as much as having a reasonable computation time.

\begin{table}
  \caption{Top1 Accuracy (\%) on FGSM attack for different $\epsilon$}
  \label{top1_FGSM}
  \centering
  \small
  \begin{tabular}{l|llllllll}
    \toprule
      & $0$ & $0.001$    & $0.005$ &  $0.01$ &  $0.05$ &  $0.1$ &  $0.15$ &  $0.5$ \\
    \midrule
    Resnet18  & 86.41 & 74.72  & 32.89 & 17.56 &  7.55 &  7.89 &  9.45 &  10.02 \\
     \midrule
    GAS-Resnet18  & \textbf{91.24} & \textbf{86.52}  & \textbf{61.12} & \textbf{41.30} & \textbf{25.27} &  \textbf{24.54}  &  \textbf{24.77} &  \textbf{15.44} \\
    \midrule
    InceptionV3  & 90.50 & 88.13  & 72.24 & 61.06 & 42.86 &  41.61  &  42.74 & 36.22 \\
  \midrule
    GAS-InceptionV3  &  \textbf{92.40} & \textbf{91.76} & \textbf{83.12} &  \textbf{71.37} & \textbf{53.51} &  \textbf{48.47}  &  \textbf{47.01} &  \textbf{40.67} \\
    \bottomrule
  \end{tabular}
\end{table}

\begin{table}
  \caption{Top1 Accuracy (\%) on PGD attack for different $\epsilon$}
  \label{top1_PGD}
  \centering
  \small
  \begin{tabular}{l|llllllll}
       & $0$ & $0.001$    & $0.005$ &  $0.01$ &  $0.05$ &  $0.1$ &  $0.15$ &  $0.5$ \\
    \midrule
    Resnet18  & 86.41 & 72.58   & 16.66 & 5.54 &  3.76 &  3.35 &  3.08 &  1.47 \\
     \midrule
    GAS-Resnet18  & \textbf{91.24} & \textbf{85.83}  & \textbf{46.72} &  \textbf{23.45}& \textbf{9.8} &  \textbf{7.16}  &  \textbf{5.5} &  \textbf{2.7} \\
    \midrule
    InceptionV3  & 90.50 & 88.13  & 61.54 &
    39.56 & 8.78  &  7.05  &  6.6 &  4.5 \\
  \midrule
    GAS-InceptionV3  &  \textbf{92.40} & \textbf{91.76}  & \textbf{79.82} & \textbf{62.42} & \textbf{16.94} &  \textbf{10.20}  &  \textbf{8.41} &  \textbf{5.3} \\
    \bottomrule
  \end{tabular}
\end{table}

\textbf{Results}. For training the GLP model, first, the GAS module was trained. The extracted global features were concatenated with local features extracted by pre-trained CNN to classify the images. 
The results in Table \ref{top1_caltech} show that the GAS model can consistently improve the performance of the CNN methods significantly. In other words, the results validate the importance of global features in the object classification task. 

\begin{table}
  \caption{Top1 Accuracy (\%) on Caltech101 dataset}
  \label{top1_caltech}
  \centering
  \begin{tabular}{llll}
    \toprule
     Resnet18 & GA-Resnet &  InceptionV3 & GA-Inception \\
    \midrule
     86.41 & \textbf{91.24} & 90.50 & \textbf{92.40}  \\
    \bottomrule
  \end{tabular}
\end{table}

Also, the the top1 accuracy results in Table \ref{top1_FGSM} and Table \ref{top1_PGD}
indicate that the GLP model is more robust than the CNNs in against both FGSM and PGD attacks overall. Also, the diagrams of the top5 and the top10 accuracy comparison are shown in Appendix \ref{app_attack} for further comparisons. 

\begin{figure}[ht]
    \centering
    \includegraphics[width=0.8\linewidth]{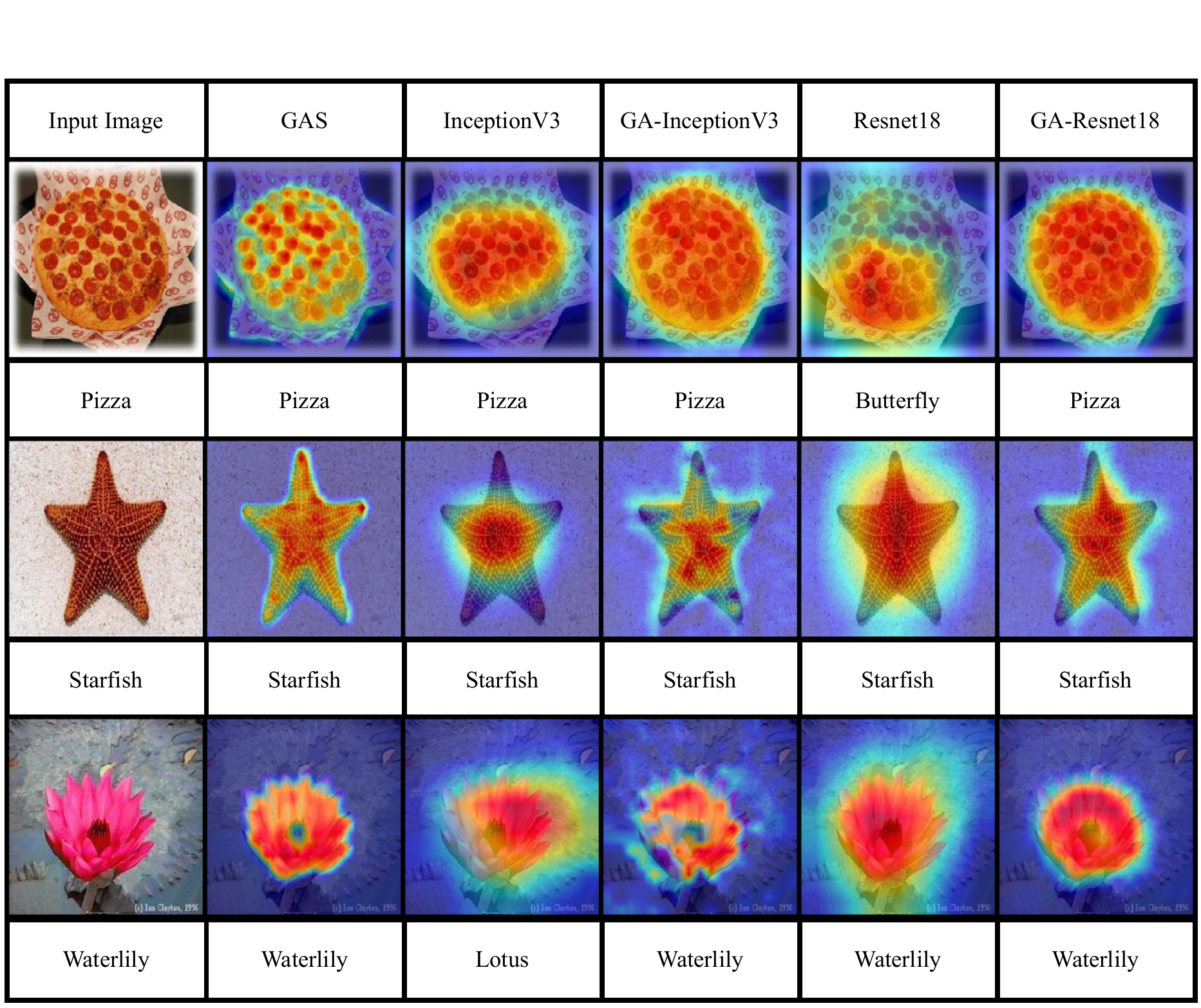}
    \caption{Comparison of visualization with Grad-Cam. In each column, the visualization results and the predicted label of the corresponding model are defined for three input images.}
    \label{fig:visualization}
\end{figure}

\textbf{Visualization} 
In order to visualize the GLP model feature maps with Grad-CAM, since this method requires a convolution layer for extracting feature maps, We provide an extra convolution layer after the GLP network and right before the concatenation operation of GLP and pre-trained CNN networks. This extra layer equalizes the size of the last convolution layer of GLP with the last convolution layer of the pre-trained network. Thus, after training this new architecture, we are capable of visualizing the feature maps of our GLP model using the Grad-CAM method. 
Fig.~  \ref{fig:visualization} depicted the comparison of Grad-CAM visualization between the CNNs and the modified version with the GAS module. The remarkable localizing objects' boundaries of the GAS network, together with the power of CNN networks, provide a more precise object localization for two-stream networks compared to using a single CNN network.

\textbf{Discussion}
In this experiment, we modified commonly used CNNs with a GAS module to improve them with an extra quick holistic view. 
According to the results, this module not only improved the accuracy of the models but also, make them more resistant to attacks. Actually, this module inspired by the subconscious function of the human pupil as an early function in perception helped these CNNs as local processors to become more holistic in the same way as humans and made the models more accurate, more robust, and more explainable.  

\section{Conclusion}
The main goal of this paper is to develop CNNs with an extra quick holistic view. To  this end, first, we introduce a new module called GAS to extract global features. The main idea behind this module is a smart filtering layer. This layer, inspired by the subconscious function of the human pupil, fades the noises using the low-pass filter. The parameter should choose smartly so that the total entropy of the whole filtered image is at its maximum. 
This new hybrid model (GLP model) has both sets of local and global features to detect images correctly. The new model not only has better performance in image classification but also the robustness of the model has increased. Also, it has been shown that the explainability of the model is improved. 
GPE, as a strong hypothesis in cognitive psychology, is not limited to the visual modality. It holds in all modalities like auditory or in language processing. So, as the feature works, it has been suggested to improve the models by adding a holistic view of other modalities and applications.  It will be hoped that these improvements not only increase the performance and robustness of the models but also helps deep learning methods  approach their main essence by imitating human at a higher level of cognition.  

\section{Compliance with Ethical Standards}
We hereby declare that there are no conflicts of interest with respect to the research presented in this paper. Furthermore, we affirm that we have not received any external funding or financial support for this research. It is important to note that this article does not include any studies involving human participants conducted by any of the authors.

\clearpage
\newpage

\medskip
\bibliographystyle{apalike}
\bibliography{ref}

\begin{thebibliography}{}

\bibitem[Aldahdooh et~al., 2021]{aldahdooh2021reveal}
Aldahdooh, A., Hamidouche, W., and Deforges, O. (2021).
\newblock Reveal of vision transformers robustness against adversarial attacks.
\newblock {\em arXiv preprint arXiv:2106.03734}.

\bibitem[Anderson et~al., 2006]{anderson2006visual}
Anderson, C.~J., Colombo, J., and Jill~Shaddy, D. (2006).
\newblock Visual scanning and pupillary responses in young children with autism
  spectrum disorder.
\newblock {\em Journal of Clinical and Experimental Neuropsychology},
  28(7):1238--1256.

\bibitem[Baker et~al., 2018]{baker2018deep}
Baker, N., Lu, H., Erlikhman, G., and Kellman, P.~J. (2018).
\newblock Deep convolutional networks do not classify based on global object
  shape.
\newblock {\em PLoS computational biology}, 14(12):e1006613.

\bibitem[Carlini and Wagner, 2017]{carlini2017towards}
Carlini, N. and Wagner, D. (2017).
\newblock Towards evaluating the robustness of neural networks.
\newblock In {\em 2017 ieee symposium on security and privacy (sp)}, pages
  39--57. IEEE.

\bibitem[de~Vries et~al., 2021]{de2021autism}
de~Vries, L., Fouquaet, I., Boets, B., Naulaers, G., and Steyaert, J. (2021).
\newblock Autism spectrum disorder and pupillometry: A systematic review and
  meta-analysis.
\newblock {\em Neuroscience \& Biobehavioral Reviews}, 120:479--508.

\bibitem[Dosovitskiy et~al., 2020]{dosovitskiy2020image}
Dosovitskiy, A., Beyer, L., Kolesnikov, A., Weissenborn, D., Zhai, X.,
  Unterthiner, T., Dehghani, M., Minderer, M., Heigold, G., Gelly, S., et~al.
  (2020).
\newblock An image is worth 16x16 words: Transformers for image recognition at
  scale.
\newblock {\em arXiv preprint arXiv:2010.11929}.

\bibitem[Fei-Fei et~al., 2004]{FeiFei2004LearningGV}
Fei-Fei, L., Fergus, R., and Perona, P. (2004).
\newblock Learning generative visual models from few training examples: An
  incremental bayesian approach tested on 101 object categories.
\newblock {\em Computer Vision and Pattern Recognition Workshop}.

\bibitem[Goodfellow et~al., 2016]{goodfellow2016deep}
Goodfellow, I., Bengio, Y., and Courville, A. (2016).
\newblock {\em Deep learning}.
\newblock MIT press.

\bibitem[Goodfellow et~al., 2014]{goodfellow2014explaining}
Goodfellow, I.~J., Shlens, J., and Szegedy, C. (2014).
\newblock Explaining and harnessing adversarial examples.
\newblock {\em arXiv preprint arXiv:1412.6572}.

\bibitem[He et~al., 2016]{he2016deep}
He, K., Zhang, X., Ren, S., and Sun, J. (2016).
\newblock Deep residual learning for image recognition.
\newblock In {\em Proceedings of the IEEE conference on computer vision and
  pattern recognition}, pages 770--778.

\bibitem[Hermann et~al., 2020]{hermann2020origins}
Hermann, K., Chen, T., and Kornblith, S. (2020).
\newblock The origins and prevalence of texture bias in convolutional neural
  networks.
\newblock {\em Advances in Neural Information Processing Systems},
  33:19000--19015.

\bibitem[Kurakin et~al., 2018]{kurakin2018adversarial}
Kurakin, A., Goodfellow, I.~J., and Bengio, S. (2018).
\newblock Adversarial examples in the physical world.
\newblock In {\em Artificial intelligence safety and security}, pages 99--112.
  Chapman and Hall/CRC.

\bibitem[Navon, 1977]{navon1977forest}
Navon, D. (1977).
\newblock Forest before trees: The precedence of global features in visual
  perception.
\newblock {\em Cognitive psychology}, 9(3):353--383.

\bibitem[Papernot et~al., 2016]{papernot2016technical}
Papernot, N., Faghri, F., Carlini, N., Goodfellow, I., Feinman, R., Kurakin,
  A., Xie, C., Sharma, Y., Brown, T., Roy, A., et~al. (2016).
\newblock Technical report on the cleverhans v2. 1.0 adversarial examples
  library.
\newblock {\em arXiv preprint arXiv:1610.00768}.

\bibitem[Paszke et~al., 2019]{paszke2019pytorch}
Paszke, A., Gross, S., Massa, F., Lerer, A., Bradbury, J., Chanan, G., Killeen,
  T., Lin, Z., Gimelshein, N., Antiga, L., et~al. (2019).
\newblock Pytorch: An imperative style, high-performance deep learning library.
\newblock {\em Advances in neural information processing systems}, 32.

\bibitem[Reddy et~al., 2020]{reddy2020biologically}
Reddy, M.~V., Banburski, A., Pant, N., and Poggio, T. (2020).
\newblock Biologically inspired mechanisms for adversarial robustness.
\newblock {\em arXiv preprint arXiv:2006.16427}.

\bibitem[Sabatino~DiCriscio et~al., 2018]{sabatino2018task}
Sabatino~DiCriscio, A., Hu, Y., and Troiani, V. (2018).
\newblock Task-induced pupil response and visual perception in adults.
\newblock {\em PloS one}, 13(12):e0209556.

\bibitem[Selvaraju et~al., 2017]{selvaraju2017grad}
Selvaraju, R.~R., Cogswell, M., Das, A., Vedantam, R., Parikh, D., and Batra,
  D. (2017).
\newblock Grad-cam: Visual explanations from deep networks via gradient-based
  localization.
\newblock In {\em Proceedings of the IEEE international conference on computer
  vision}, pages 618--626.

\bibitem[Szegedy et~al., 2015]{szegedy2015going}
Szegedy, C., Liu, W., Jia, Y., Sermanet, P., Reed, S., Anguelov, D., Erhan, D.,
  Vanhoucke, V., and Rabinovich, A. (2015).
\newblock Going deeper with convolutions.
\newblock In {\em Proceedings of the IEEE conference on computer vision and
  pattern recognition}, pages 1--9.

\bibitem[Turi et~al., 2018]{turi2018pupillometry}
Turi, M., Burr, D.~C., and Binda, P. (2018).
\newblock Pupillometry reveals perceptual differences that are tightly linked
  to autistic traits in typical adults.
\newblock {\em Elife}, 7:e32399.

\bibitem[Vaswani et~al., 2017]{vaswani2017attention}
Vaswani, A., Shazeer, N., Parmar, N., Uszkoreit, J., Jones, L., Gomez, A.~N.,
  Kaiser, {\L}., and Polosukhin, I. (2017).
\newblock Attention is all you need.
\newblock {\em Advances in neural information processing systems}, 30.

\bibitem[Wandell, 1995]{wandell1995foundations}
Wandell, B.~A. (1995).
\newblock {\em Foundations of vision.}
\newblock Sinauer Press,.

\end{thebibliography}

\clearpage

\section*{Appendix}
\appendix

\section{Performance Comparison}
\label{model_performance}
Inference time (ms/image), number of model's parameters (M), and model sizes (MB) for different input image sizes according to the standard input sizes; $224\times224$ for Resnet18 and $299\times299$ for the Inceptionv3 were reported in Table \ref{app_performance}. 

\begin{table}
  \caption{Comparison of models' performances}
  \label{app_performance}
  \centering
  \begin{tabular}{llllll}
    \toprule
     Models & Inference Time (ms/img) & Params (M) & Model Size (MB) & Input-size \\
    \midrule
     GAS-299 & 1.7 & 17.9 & 71.6 &  $299\times299$\\
    \midrule
     InceptionV3 & 3.8 & 24.6 & 98.9  &$299\times299$\\
    \midrule 
      GA-Inception & 4.3 & 75.3 & 301.6 &  $299\times299$\\
      \midrule
     GAS-224  & 1.1  & 10.3  &41.4 & $224\times224$\\
    \midrule
    Resnet18  &  1.4 & 11.2 & 45.5 & $224\times224$\\
    \midrule
    GA-Resnet & 1.6 & 34.1 & 136.9 & $224\times224$\\

    \bottomrule
  \end{tabular}
\end{table}

\section{Adversarial Attacks Comparisons}
\label{app_attack}
\begin{figure}[ht]
    \centering  \includegraphics[width=0.8\linewidth]{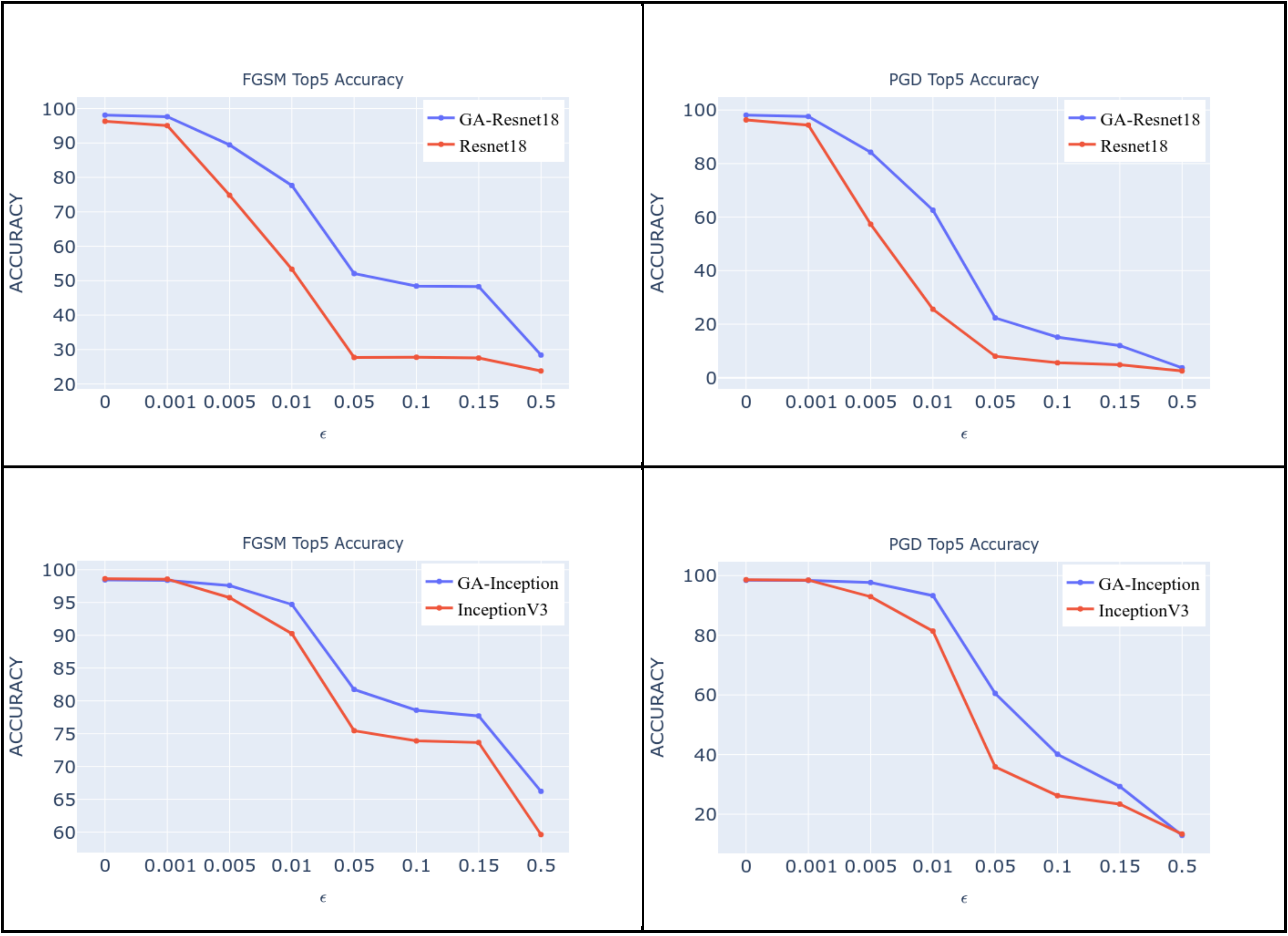}
    \caption{Comparison of Top5 accuracy on FGSM and PGD attacks}
    \label{fig:top5_attack}
\end{figure}

\begin{figure}[ht]
    \centering
    \includegraphics[width=0.8\linewidth]{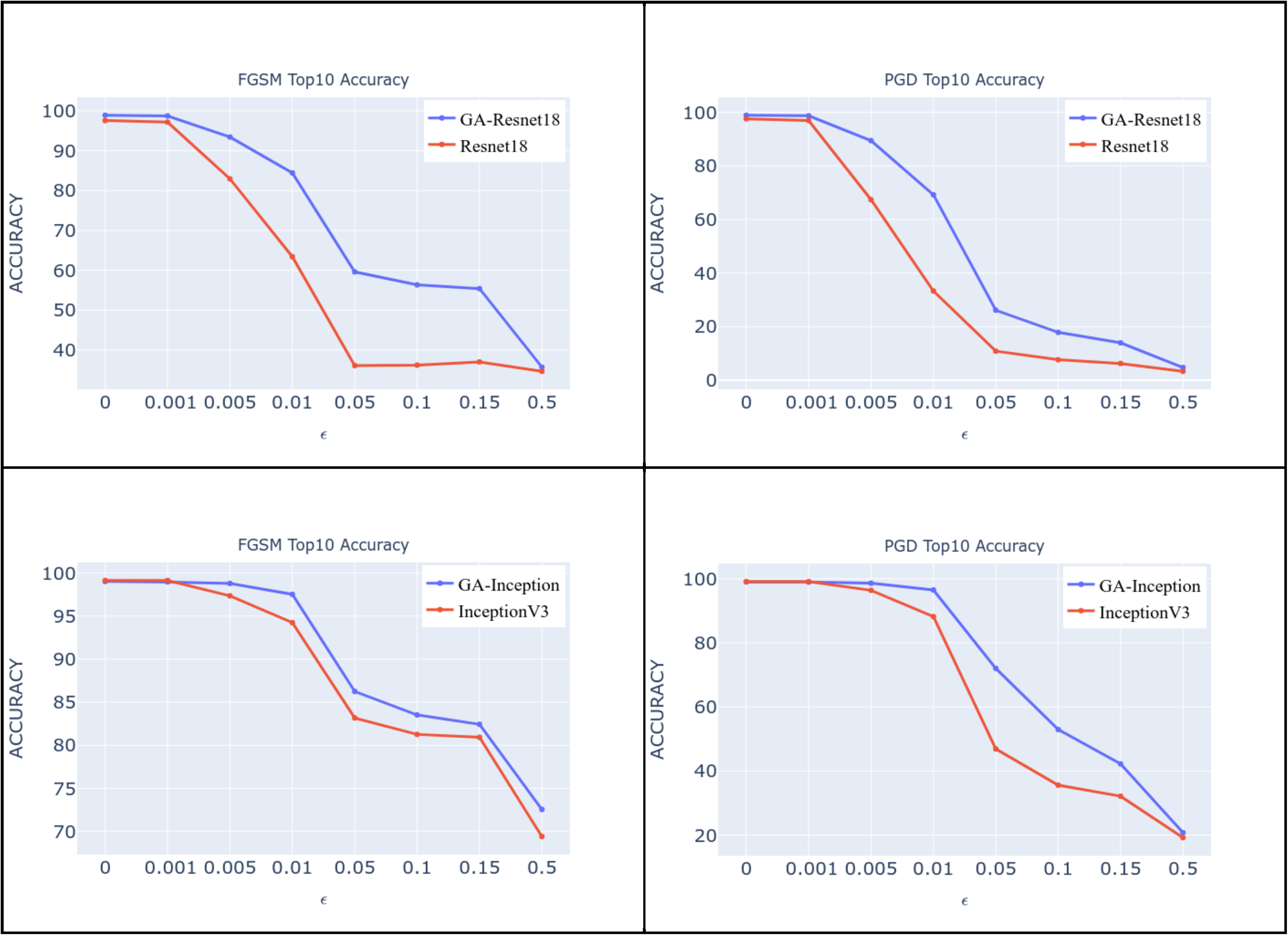}
    \caption{Comparison of Top10 accuracy on FGSM and PGD attacks}
    \label{fig:top10_attack}
\end{figure}

The top5 and top10 accuracy diagrams are depicted in  \ref{fig:top5_attack} and \ref{fig:top10_attack} figures, respectively. The results confirm that the GLP model preserves its robustness against both FGSM and PGD attacks than the CNNs.

\end{document}